\documentclass[12pt]{article}
\usepackage{amssymb,amsmath,amsthm}

\theoremstyle{plain}

\begin{document}
\title{On the Limits of Hierarchically Embedded Logic in Classical Neural Networks}
\author{Bill Cochran}
\maketitle

\begin{abstract}
We propose a formal model of reasoning limitations in large neural net models for language,
grounded in the depth of their neural architecture.
By treating neural networks as linear operators over logic predicate space we 
show that each layer can encode at most 
one additional level of logical reasoning. Logic classes $\mathcal{L}_k$ are 
defined inductively, with $\mathcal{L}_0$ representing atomic predicates and 
$\mathcal{L}_{k+1}$ formed via quantification or composition over $\mathcal{L}_k$. 
We prove that a neural network of depth $n$ cannot faithfully represent 
predicates in $\mathcal{L}_{n+1}$, such as simple counting over complex 
predicates, implying a strict upper bound on logical expressiveness. This structure 
induces a nontrivial null space during tokenization and embedding, excluding 
higher-order predicates from representability. Our framework offers a natural 
explanation for phenomena such as hallucination, repetition, and limited planning,
while also providing a foundation for understanding how 
approximations to higher-order logic may emerge. These results motivate architectural 
extensions and interpretability strategies in future development of language
models.
\end{abstract}

\section{Introduction}
Language is tough to model. This paper proposes a linear construction
of the bit-level structure of a nonlinear neural network designed for language,
and demonstrates that these bits are incapable of distinguishing objects
whose complexity exceeds the number of bits in the model. This limitation
arises from the sheer number of logical predicates required
to describe such objects.

This construction enables the interpretation of logic predicates
as linear combinations of lower-order predicates, providing
a framework for interpolation: the nodes of the language model
represent linear combinations of known algebraic predicates. These
combinations provide local support for meaning interpolation.

Through this lens, we can characterize the types of predicates
excluded by the model’s null space---for example, those needed 
to distinguish highly complex objects---and develop a linear model
that can be used to understand the system’s nonlinearity. This would 
improve compression, enhance neural network efficiency, and support 
more accurate interpolation of intelligence.

The model also provides a framework for understanding
the origin of nonlinearity in neural networks trained on language.
Each predicate is a linear sample from an interpolation space of lower-order predicates.
Higher-order predicates are evaluated using only an interpolated subset of lower-order logic.
This incomplete evaluation leads the model to construct necessarily incomplete---
and therefore potentially incorrect---predicates.

\section{Related Work}
This paper develops a model of language token generation rooted in classical
approximation theory.  We break the process into logical parts, each having
been studied quite exstensively and describe operators that perform identical
operations. We describe these tensors conversationally here and mathematically
below.

We begin with the abstraction of meaning into tokens: a non-linear mapping 
between natural language and a finite symbolic space. While such tokenizations 
are essential for computability, they introduce a null space — compressing 
distinctions that may be undecidable, logically subtle, or simply orthogonal 
to the learned basis. This view generalizes the insight of Shannon, who first 
formalized communication as a lossy channel, and is sharpened by modern 
critiques of token-level semantics. We accomplish this mapping with two tensors.

Our tensor $T$ reflects this mapping as an abstraction of meaning, some
semantic unit is tagged with a unique natural number. The second tensor,
$L^+$, represents structural compression, one whose limitations propagate through 
every downstream operation \cite{shannon1948, bender2020climbing}. 
To wit: $T$ abstracts meaning from tokens and $L^+$ compresses these
tokens.

The function of $L^+$ — mapping from token space to discrete logic — parallels 
classical interpolation theory. Weierstrass proved that all continuous functions 
may be approximated by polynomials, yet as later formalized by Cheney and others, 
such approximations can introduce substantial error when the underlying basis is 
misaligned with the target function. Language models, in this light, act as 
interpolation machines operating over an incomplete logical basis. The resulting 
aliasing — structurally indistinguishable projections of semantically distinct 
inputs — is a numerical inevitability, not a design flaw \cite{cheney1966approx, 
courant1924methoden}.

The final tensor we assume, $L$, completes the circuit by forward projecting selection 
logits onto predicate space---reconstructing a representation of meaning 
from utility. This projection is necessarily underspecified: for a given activation 
pattern, many predicate configurations may be plausible. As Hadamard first 
formalized, such inverse mappings are ill-posed \cite{hadamard1902problemes}, 
requiring regularization or prior structure to produce stable interpretations 
\cite{tikhonov1963regularization}. The phenomenon of hallucination in language 
models thus emerges not from failure, but from the ambiguity inherent in 
decoding underdetermined logical structure.

This work also builds on classical theories of logical expressivity and recent studies on the representational power of neural networks:

\begin{itemize}
\item Fagin~\cite{fagin1974generalized} and Immerman~\cite{immerman1999descriptive} formalized the hierarchy of logical predicates in connection with computational complexity.
\item Telgarsky~\cite{telgarsky2016benefits} and Hanin~\cite{hanin2019complexity} demonstrate that deeper neural networks can represent functions that shallow ones provably cannot.
\item Weiss et al.\cite{weiss2021thinking} and Saxton et al.\cite{saxton2019analysing} explore the limitations of large language models in symbolic and arithmetic reasoning.
\item Lanham and Nikolić~\cite{lanham2023logical} introduce a notion of logical depth in language models, conceptually aligned with our tensor formalism.
\end{itemize}

We complement these insights by offering a tensor-based formalization of 
semantic unit selection in language models, and by proving a bound on i
logical expressiveness as a function of depth. A full review will appear 
in future versions. 

\section{Bound on Distinguishability In A Neural Net}
Let $\mathcal{L}_i$ be the class of $i$-th order logic. Let $N_{d,w}$ 
describe a neural network with $d$ layers and $w$ nodes per layer. We 
demonstrate:

\begin{quote}
For all $w \in \mathbb{N}$, there exists a predicate $p \in \mathcal{L}_2$ 
that cannot be represented by any one-layer network $N_{1,w}$. In particular:
\end{quote}
\begin{quote}
\emph{There are at least $w+1$ true distinct propositions can be implied from the input.}
\end{quote}

This predicate is in $\mathcal{L}_2$: 
$$
    \exists i_1, \dots, i_{w+1} \in \{1, \dots, n\} \quad
    \left(
    \bigwedge_{1 \leq j < k \leq w+1} i_j \neq i_k \ \wedge\
    \bigwedge_{j=1}^{w+1} P(i_j)
    \right).
$$
It cannot be represented by $N_{1,w}$, which can linearly combine at most 
$w$ predicates. Thus, such threshold predicates are in the network's null 
space.  From here, it is trivial to construct a countably infinite set
of decidable predicates that cannot be determined.  Trivially, this neural
net can only distinguish $w$ atoms (or appropriately sized linear combinations
of).  Within this framework, a neural net approximates higher order logic
by having a finite and fixed set predicates that determine its value.
So, the network can also distinghuish bounded linear combinations of $w$
things, give or take the atoms necessary to abstract orders of logic.
For example, instead of $\forall$, the network can imitate any appropriately
bounded finite state automaton.  The bound is computable from $w$ through
induction and demonstrates how this low order neural net can produce
high order logic on any appropriately sized finite set of tokens.

We then define an inductive construction: assume $P_{d+1} \subset 
\mathcal{L}_{d+1}$ is a countably infinite set of predicates requiring depth 
$d+1$ that cannot be determined by net $N_{d,w}$. Then define:
\begin{quote}
\emph{``There are at least $d \times w + 1$ true distinct predicates in $P_{d+1}$ implied by the input.''}
\end{quote}
Restating in predicate logic:
$$
    \exists p_1, \dots, p_{d \times w + 1} \in P_{d+1} \quad
    \left(
    \bigwedge_{1 \leq j < k \leq d \times w + 1} p_j \neq p_k \ \wedge\
    \bigwedge_{j=1}^{d \times w + 1} p_j(\text{input})
    \right)
$$
For these predicates, if we choose from the countable set of predicates that
are unrepresentable by the network, $P_{d+1}$, then this predicate is not
just undecidable, but the network has no information to decide it. The
countably infinite constructor trivially follows.

Through precise construction and enumeration, one can exhaust the logical capacity
of any neural network model by demanding it count beyond its representational limits.
In order to make this neural network model linear, we define $\hat{\mathcal{T}}$ as 
the token set of a language model: the set of atoms the neural network can reason about. 
Necessarily,
$\hat{\mathcal{T}}\subset \mathcal{P}$, the set of all decidable predicates.  In short,
a predicate $p\in \mathcal{P}$ but $p \notin \hat{\mathcal{T}}$, has been
demonstrated to exist.  This predicate describes an appropriately complex object. 
In short, it is possible to compute an object that a neural net cannot reason about
just from the number of nodes in the network.

In particular, a finite counting of complicated finite objects is enough to exhaust
any finite neural net.  There exist two objects complex enough that the neural net
cannot distinguish, no matter how it was trained.  This is a common sense result
and underscores that reasoning seen otherwise is not error free.  This is
different behavior than most computer algorithms, which tend to fail in 
exploitable ways when reasoning is not to be trusted.

\section{Constructing the Neural Net from Prediate Logic}

By construction, each component of a neural network---weights, 
activations, and transformations---can be expressed as logical 
predicates. For a fixed architecture $N_{d,w}$, once its 
structure and parameters are specified, its behavior on any 
given input is fully determined and thus decidable.

Therefore, $N_{d,w}$ defines a complex predicate:
$$
    \mathsf{Net}_{d,w}(x, y) := \text{``The network } N_{d,w} \text{ produces output } y \text{ given input } x." 
$$
Every input deterministically produces a single activation 
pattern across the network. This makes the entire computation 
trace a compound logical predicate---essentially, a chain of 
nested implications that resolve to a binary outcome. Put 
plainly: once you know the wiring and weights, the net can’t 
do anything else.

Even individual nodal values can be recovered bit by bit in
predicate space.
\begin{quote}
    The $i$-th bit of the $j$-th node of neural net $k$ is 1 given this input.
\end{quote}

Ergo, we have built a high dimensional linear operator that encompasses
every possible internal state any simply connected, dense neural net can be in.

By linearizing the neural net, we can identify how and why
complex predicates are modeled within the neural network.  We present
a tensor construction that explores how the neural net evolves in
an idealized natural langugate token generation process.

\section{The Null Space in Language Models}
We decompose the information encoding/decoding steps of language processing
using four tensors:
\begin{enumerate}
\item[$T$] \emph{Tokenizer} We assume an idealized, non-linear mapping 
 between natural language and an infinite token space. In practice, this 
 mapping has a nontrivial null space: certain linguistic distinctions or 
undecidable propositions may be irrecoverably compressed, rendering them 
unrepresentable in token space.  These propositions are left as an
        exercise for the reader.
    \item[$L^+$] \emph{Backward Logit Projection} This operator maps all 
        possible tokens in natural language to the set of 
predicates whose truth values the language model can determine. Due to the model’s 
limited logical depth, $L^+$ has a nontrivial null space consisting of predicates 
that cannot be represented. This results in aliasing: multiple distinct concepts may 
project to the same representation, allowing fluent language output despite an 
incomplete reasoning foundation. We can use the Moore-Penrose or other minimum residual inverse.
    \item[$M$] \emph{Aribtrary Neural Net} This is the one-to-one mapping computed non-linearly by the neural net, 
    but linearly by predicate logic.  We constructed this operator to be invertible.
\item[$S$] \emph{Logit Selection Operator} This one-to-one mapping takes an activation and computes some utility for 
    token selection.  We assume this operator is invertible.
\item[$L$] \emph{Forward Logit Projection}  This operator converts logit selection criteria to the
    infinite space of predicates. This is an underspecified projection, as discussed above.
\item[$T^{-1}$] We assume that the tensor in step 1 is invertible
\end{enumerate}
In particular, each of these operators work on $\mathbb{N}^{x\times y}$ identifiers
and return a $\mathbb{N}^{i\times j}$, for $i,j,x,y\in\mathbb{N}$.  These numbers are tags 
that substitute for predicates, semantic units, and other less precise terms.

A process for next semantic unit generation of language might be.
$$
p_{i+1} = T^{-1} L S M  L^+ T (p_i,p_{i-1},\ldots)
$$

Routine numerical analyses of these operators describe the interpolation process of 
information retrieval and abstraction.  For instance, $L$ is a highly underspecified
projection, where a large amount of information can be interpolated from the 
associated null space of $L^+$.

Let's consider the information in the null space of $L^+$.  The construction of the
finite neural net logic result suggests that there is an interpolation of higher
order logic that can take place.  This interpolation can be seen as a mask
applied to the complete linear combination of whatever logic predicate is being
used as an interpolant for the next token.

This suggests that the interpolation samples the actual true predicates and estimates
truth of others by this interpolation.  The estimation of truth is more than likely
shared across various concepts, related and unrelated.  In other words, the predicates
being selected from are "generally" true or "generally" false because this logic
has appeared in other contexts.  The generally in this case, is probably associated
with a minimum residual calculation of some sort.

So some $p\in \mathcal{L}_b$ b-order logic predicate can be approximated by a finite set 
$n_i\in\mathcal{L}_d$, $b \gg n$,  order finite approximates in the mesh.
$$
p\approx\sum_i \alpha n_i
$$
It is this approximation that loses information and is the source of hallucination in
this model.  
I refer to this equation as the \emph{metaphor metaphor}:
it highlights that certain aspects of a higher-order predicate can be preserved---or at least
evoked---by a weighted combination of lower-order logic. Often, the result is
structurally useful but not logically precise---like a metaphor.

\section{Acknowledgements}

This work was developed in collaboration with ChatGPT-4o and 4.5. ChatGPT was 
able to engage with and extend this model in compelling ways, often with surprising 
fluency. As it aptly summarized the source of hallucinations:

\begin{quote}
It’s recovering a low-rank projection of something that was never full-rank to begin with.
\end{quote}

\bibliographystyle{plain}
\bibliography{logic}
\end{document}